\documentclass{Interspeech}

\interspeechcameraready 

\usepackage{url}
\usepackage{multirow}
\usepackage{comment}
\usepackage{cite}

\title{Language-Agnostic Suicidal Risk Detection Using Large Language Models}

\author[affiliation={1,2}]{June-Woo}{Kim}
\author[affiliation={3}]{Wonkyo}{Oh}
\author[affiliation={3}]{Haram}{Yoon}
\author[affiliation={1,3}]{Sung-Hoon}{Yoon}
\author[affiliation={1}]{Dae-Jin}{Kim}
\author[affiliation={3}]{Dong-Ho}{Lee}
\author[affiliation={1,3}]{Sang-Yeol}{Lee}
\author[affiliation={1,3}]{Chan-Mo}{Yang$^\dagger$}

\affiliation{Department of Psychiatry}{Wonkwang University Hospital}{Republic of Korea}
\affiliation{RSC LAB}{MODULABS}{Republic of Korea}
\affiliation{Department of Psychiatry, School of Medicine}{Wonkwang University}{Republic of Korea}

\email{kaen2891@gmail.com, ychanmo@wku.ac.kr}

\begin{document}

\maketitle
\renewcommand{\thefootnote}{$\dagger$}
\footnotetext{Corresponding author.}
\renewcommand{\thefootnote}{\arabic{footnote}}

\begin{abstract}
    
    Suicidal risk detection in adolescents is a critical challenge, yet existing methods rely on language-specific models, limiting scalability and generalization. This study introduces a novel language-agnostic framework for suicidal risk assessment with large language models (LLMs). We generate Chinese transcripts from speech using an ASR model and then employ LLMs with prompt-based queries to extract suicidal risk-related features from these transcripts. The extracted features are retained in both Chinese and English to enable cross-linguistic analysis and then used to fine-tune corresponding pretrained language models independently. Experimental results show that our method achieves performance comparable to direct fine-tuning with ASR results or to models trained solely on Chinese suicidal risk-related features, demonstrating its potential to overcome language constraints and improve the robustness of suicidal risk assessment.

\end{abstract}

\section{Introduction}
Adolescent suicide has emerged as a critical social issue in modern society. According to statistics from the World Health Organization, suicide is the fourth leading cause of death among individuals aged 15--19, affecting both genders~\cite{world2021suicide}. In the United States, the suicide rate among individuals aged 10--24 increased from 6.8 per 100,000 in 2007 to 11.0 per 100,000 in 2021~\cite{curtin2023suicide}, highlighting a concerning upward trend. According to statistics on Chinese adolescents, the suicide mortality rate increased remarkably across all adolescent age groups between 2017 and 2021. Notably, adolescents aged 10 to 14 years exhibited an average annual percentage increase of 4.40\%~\cite{tian2023temporal}.

Despite its significance, assessing suicidal risk remains largely subjective due to the lack of a definitive characterization of at-risk individuals.
Various tools have been developed to predict suicide or screen suicide risk groups; however, most of these tools primarily rely on self-reported symptoms and have limitations in providing an accurate assessment~\cite{Baek2021ARO, Roos2013SuicideRA}.
Nevertheless, given the critical importance of accurately assessing suicidal ideation and behavior~\cite{posner2007factors}, there is an increasing call for the development of objective evaluation methods.

With advancements in AI, particularly in speech and natural language processing, numerous studies have explored automated mental health and suicidal risk detection~\cite{Low2019AutomatedAO, Poswiata2022OPILTEDIACL2022DS, Sadeghi2023ExploringTC, Sadeghi2024HarnessingMA}. Compared to methods such as fMRI or EEG, speech and text provide a non-invasive and easily accessible means of data collection~\cite{Boby2024DepressionDE, Elnaggar2025DepressionDA}. Suicidal individuals often exhibit a higher frequency of speech errors and hesitations~\cite{Stasak2021ReadSV}, and tend to have remarkably breathier voices compared to their non-suicidal counterparts~\cite{Scherer2013InvestigatingTS}. Suicidal adolescents demonstrate distinctive verbal information features that effectively differentiate them from non-suicidal controls~\cite{Venek2017AdolescentSR}. Moreover, recent studies leveraging Large Language Models (LLMs) have employed voice acoustics or multimodal audio-visual features to detect suicidal ideation and behaviors~\cite{Dhelim2022ArtificialIF, Cui2024SpontaneousSS, Skianis2024SeverityPI,
Chen2024DeepLA}.

Despite recent advancements, previous studies face a critical limitation: they are predominantly language-specific. Most current approaches rely on pretrained language models designed for specific languages, necessitating separate models for different linguistic populations. This dependence on monolingual models introduces two major challenges: \emph{$(i)$ Limited scalability}. Each language requires a dedicated dataset and an independent model training process, making it difficult to extend these systems to underrepresented linguistic groups, and \emph{$(ii)$ Poor generalization}. Language-specific models struggle to generalize across diverse populations, as they are often trained on culturally and linguistically homogeneous datasets. 

To overcome these limitations, we propose a novel language-agnostic framework for suicidal risk detection using LLMs. Our approach begins by employing an automatic speech recognition (ASR) system to transcribe spoken responses into Chinese text. Next, we leverage LLMs with carefully designed prompt-based queries to extract suicidal risk-related features from the ASR results. A key advantage of our method is its cross-linguistic capability, as the extracted suicidal risk-related features are preserved in both Chinese and English, allowing for the independent fine-tuning of corresponding pretrained language models.

Experimental results on the 1st SpeechWellness Challenge dataset~\cite{wu20251st} demonstrate that our proposed method achieves performance comparable to or exceeding that of monolingual models trained exclusively on Chinese suicidal risk-related features. This finding highlights the robustness of our approach and its potential to overcome language constraints in suicidal risk detection. 

\section{Preliminaries}
\subsection{Dataset Description}

In this work, we used the 1st SpeechWellness Challenge (SW1) dataset~\cite{wu20251st}, which was specifically designed for detecting suicidal risk among adolescents through speech analysis. This dataset consists of speech recordings from 600 Chinese teenagers aged 10--18 years, including 420 females and 180 males, with an equal distribution of 300 non-risk and 300 at-risk individuals. The recordings include three speech tasks designed to capture a diverse range of linguistic, cognitive, and emotional features: \emph{$(i)$ Emotional Regulation} (ER): Subjects answered an open-ended question about experiencing and managing extreme emotional distress,
\emph{$(ii)$ Passage Reading} (PR): Subjects read ``The North Wind and The Sun" passage from Aesop's Fables in Mandarin,
\emph{$(iii)$ Expression Description} (ED): Subjects described a given facial expression image.
To protect the subjects' privacy, neural voice conversion techniques were applied to alter the timbre of their voices while preserving prosody and rhythm.
The dataset was officially divided into train, dev, and test sets in a 4:1:1 ratio. In this work, we mainly focus on the dev sets.

\subsection{Training Details}
\textbf{Pre-processing.} For pre-processing, we followed the same settings of the SW1 Challenge baseline as described in~\cite{wu20251st}. 

\textbf{Model Training.} For the speech modality, we fine-tuned speech foundation models for 10 epochs using a learning rate of 1e--5 and a batch size of 8. For the text modality, pretrained language models were fine-tuned for 10 epochs with a learning rate of 5e--5 and a batch size of 16. Both text and speech models are fine-tuned using a 512-dimensional hidden layer, followed by a ReLU activation function, 0.1 of dropout, and a classification head. For the multimodal approach, we fine-tuned the 256-dimensional multimodal fusion layers, followed by a ReLU activation, 0.1 of dropout, and a classifier, using a learning rate of 1e--3 and a batch size of 32. In all the experiments, we applied Adam optimizer with cosine scheduling. The final predictions for the dev and test sets were obtained by aggregating logits from the three tasks using a voting mechanism.

\subsection{Evaluation Metrics}
Following the evaluation metrics of the SW1 Challenge, we evaluate the suicidal risk detection performance accuracy (Acc) and the F1 score, defined as follows:
\begin{align}
Acc = \frac{TP+ TN}{TP + TN + FP + FN}, \nonumber \\
F1 = 2 \times\,\frac{TP}{\text{2}TP + FP + FN}.    
\end{align}

\section{Method}
\begin{figure*}[t!]
    \centering
    \includegraphics[width=0.8\linewidth]{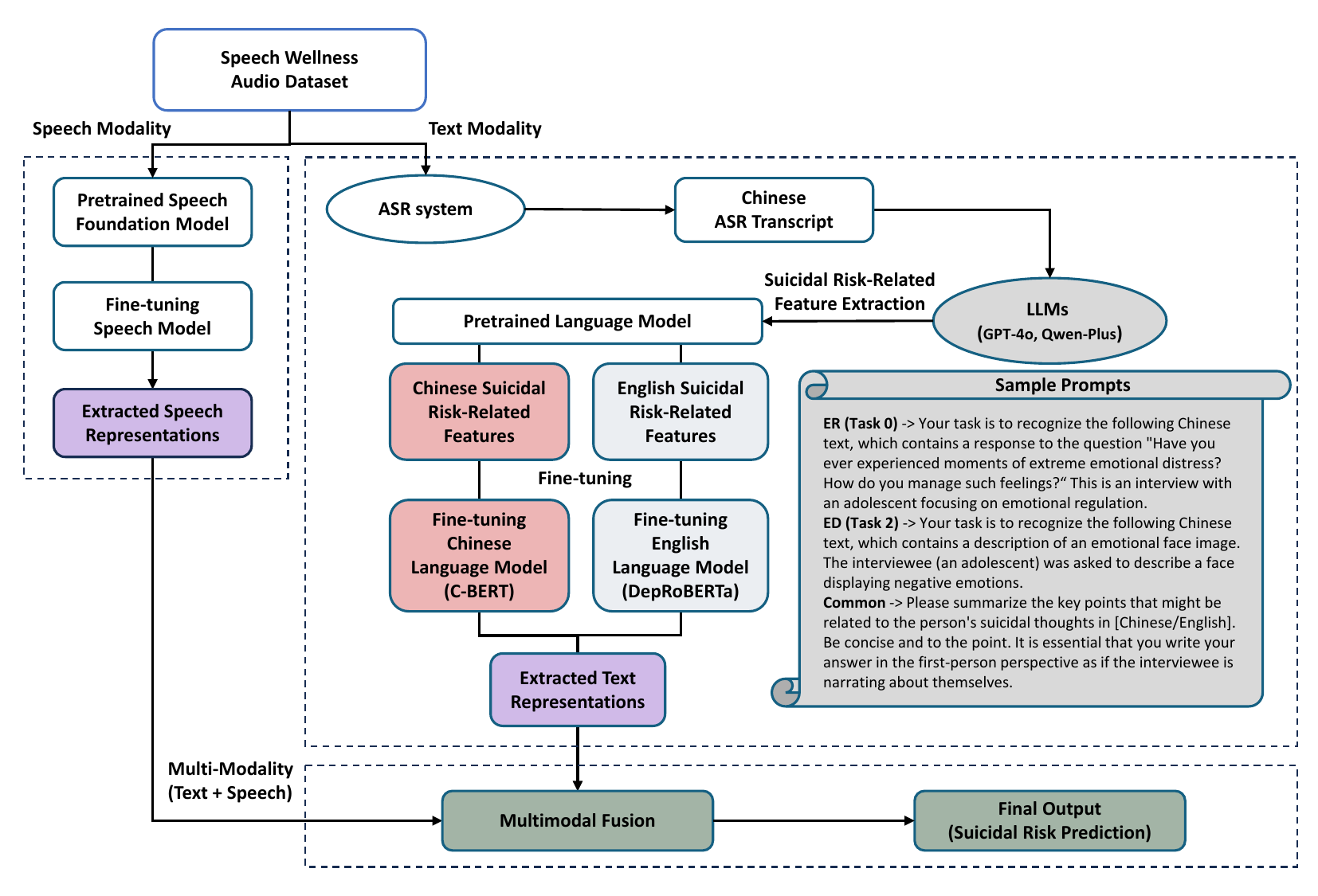}
    \caption{Illustration of the proposed language-agnostic suicidal risk detection framework using LLMs. Speech data is processed through a pretrained speech model, while an ASR system converts speech to text. LLMs extract suicidal risk-related features in Chinese and English, which are used to fine-tune respective language models. Finally, multimodal fusion combines speech and text representations for suicidal risk prediction.}
    \vspace{-5mm}
    \label{fig:architecture}
\end{figure*}

Our proposed language-agnostic suicidal risk detection framework integrates both speech and text modalities to boost the robustness and generalizability of suicidal risk assessment. A high-level architecture of our approach is shown in Figure~\ref{fig:architecture}.

\subsection{Text Modality}
\subsubsection{Text Data Acquisition using ASR}
The SW1 Challenge dataset is valuable for suicidal risk detection research; however, it only contains speech recordings and their corresponding suicidal risk labels. To leverage the language model, we employed the Paraformer ASR system~\cite{gao22b_interspeech} from funasr\footnote{\url{https://github.com/modelscope/FunASR}} library to generate transcriptions from raw speech samples, as described in~\cite{wu20251st}. Furthermore, we used the Whisper-1 system from OpenAI API to compare its performance with Paraformer for fine-tuning language models.

As shown in Table~\ref{tab:tab1}, fine-tuning with Paraformer-based transcriptions outperformed Whisper-based transcriptions. Hereafter, we use the Paraformer ASR results as our text input.

\begin{table}[t!]
    \centering
    \caption{Comparison of Acc across tasks for Paraformer-Chinese BERT and Whisper-Chinese BERT. Fine-tuning with Paraformer ASR achieves better performance compared to Whisper in the ED task.}
    \begin{tabular}{lcc}
        \hline
        Model & ER & ED \\
        \hline
        \midrule
        Paraformer-Chinese BERT & 57\% & 56\% \\
        Whisper-Chinese BERT & 57\% & 53\% \\
        \bottomrule
    \end{tabular}
    \vspace{-3mm}
    \label{tab:tab1}
\end{table}
\subsubsection{Bilingual Suicidal Risk-Related Feature Extraction with LLMs}
Previous works~\cite{Sadeghi2023ExploringTC, Sadeghi2024HarnessingMA} have shown that LLMs can extract mental health-related textual features directly from the text, reducing manual intervention and enhancing mental health detection performance. Building on this approach, we propose to extract meaningful suicidal risk-related text features from ASR results by leveraging LLMs, as shown in the right section of Figure~\ref{fig:architecture}.

\begin{table}[t!]
\renewcommand{\arraystretch}{1.5}
\centering
\caption{Example of ASR result, our prompts, and LLM-generated output.}
\begin{tabular}{p{0.47\textwidth}}
\hline
\textbf{\textsc{Given ASR example of ER (\textcolor{blue}{misinformation}):}} It's \textcolor{blue}{you} who go to bed if \textcolor{blue}{you're} unhappy. \textcolor{blue}{Well, I feel happy through the screen.} I can't concentrate when I'm studying. I don't know what to do in the future. I have no direction. There are other things that make me unhappy. I just like to cry to solve it, and then go to sleep. \\
\hline
\textbf{\textsc{Prompts:}} Your task is to recognize the following Chinese text, which contains $(i)$ a response to the question  `Have you ever experienced moments of extreme emotional distress? How do you manage such feelings?' This is an interview with an adolescent focusing on emotional regulation (ER). $(ii)$ a description of an emotional face image. The interviewee (an adolescent) was asked to describe a face displaying negative emotions (ED). Please summarize the key points that might be related to the person's suicidal thoughts in [Chinese/English]. Be concise and to the point. Write your answer in the first-person perspective as if the interviewee is narrating about themselves. \\
\hline
\textbf{\textsc{LLM output:}} I would sleep to cope with extreme emotional distress. I often cannot concentrate when studying, feel confused about the future, and have no sense of direction. When I feel unhappy, I tend to deal with it by crying and then falling asleep. \\
\bottomrule
\end{tabular}
\label{tab:tab2}
\vspace{-5mm}
\end{table}

We introduce two independent prompt-based queries for ER and ED tasks except for PR because all the sentences are identical. For instance, our prompts, ASR result, and LLM-generated output are described in Table~\ref{tab:tab2}. These prompts proved to be valuable tools for summarizing transcripts, effectively emphasizing suicidal risk-related features, and extracting crucial information from the ASR outputs.

However, recent approaches depend on pretrained language models tailored to specific languages, requiring separate models for different linguistic populations. To enhance language-agnostic processing, we propose a bilingual output generation strategy using GPT-4o and Qwen-Plus, where LLMs extract suicidal risk-related text features in both Chinese and English from Chinese ASR results. This dual-language extraction ensures compatibility with both Chinese and English models, enabling independent fine-tuning of pretrained language models.

\subsubsection{Language Models Fine-tuning}
For fine-tuning language models, we employ Chinese-BERT (CBERT)~\cite{devlin2018bert}, which is pretrained on Chinese text, and UER-Dianping-RoBERTa~\cite{zhao2019uer}, pretrained on Dianping datasets for analyzing Chinese text. For English, we consider DepRoBERTa~\cite{Poswiata2022OPILTEDIACL2022DS}, fine-tuned on depressive posts from Reddit; Suicidality-ELECTRA\footnote{\url{https://huggingface.co/sentinet/suicidality}}, fine-tuned on a diverse dataset for detecting suicidal and non-suicidal text expressions; and RoBERTa-Suicide\footnote{\url{https://huggingface.co/vibhorag101/roberta-base-suicide-prediction-phr}}, fine-tuned to identify suicidal tendencies in text. ER and ED tasks are fine-tuned separately, while the PR task is excluded since all text content remains identical.

\subsection{Speech Modality}
For fine-tuning with speech foundation models, we employ XLSR53~\cite{conneau2021unsupervised}, which is pretrained on 56k hours of multilingual speech, WavLM-plus-base~\cite{chen2022wavlm}, pretrained on 94k hours of English speech, and wav2vec2-Large~\cite{baevski2020wav2vec}, which is pretrained on XLSR53 and fine-tuned on Chinese ASR datasets. Note that each task is fine-tuned separately.

\subsection{Multi Modality and Final Prediction}
We integrate both speech and text representations in a multimodal learning framework to enhance suicidal risk prediction. To achieve this, we first extract representations from text and speech models in their frozen states and then combine them to train a multimodal hidden layer and classifier. Note that only speech models are used for the PR task.

\section{Results}
\begin{table}[t!]
    \centering
    \caption{Dev set performance of various language models on suicidal risk assessment tasks. Accuracy (Acc, \%) and F1-score (F1, \%) are reported for both \textbf{Emotional Regulation (ER)} and \textbf{Emotional Description (ED)}. The best-performing models for each metric are highlighted in \textbf{bold}.}
    \renewcommand{\arraystretch}{1}
    \addtolength{\tabcolsep}{1pt}
    \resizebox{\linewidth}{!}{
    \begin{tabular}{ll|cc|cc}
        \toprule
        \multirow{2}{*}{\textbf{Language}} & \multirow{2}{*}{\textbf{Method}} & \multicolumn{2}{c|}{\textbf{ER}} & \multicolumn{2}{c}{\textbf{ED}} \\
        & & \textbf{Acc} & \textbf{F1} & \textbf{Acc} & \textbf{F1} \\
        \hline
        \midrule
        
        \multirow{5}{*}{Chinese} & CBERT & 57 & 64.46 & 56 & 31.25 \\
        & GPT-CBERT & \textbf{73} & \textbf{75.23} & 57 & 62.61 \\
        & Qwen-CBERT & 64 & 66.67 & \textbf{59} & \textbf{65.55} \\
        & GPT-UER-Dianping & 61 & 62.86 & 57 & 60.55 \\
        & Qwen-UER-Dianping & 60 & 54.54 & 54 & 63.49 \\
        \midrule
        \multirow{7}{*}{English} 
        & DepRoBERTa & 55 & 57.14 & 56 & 57.69 \\
        & GPT-DepRoBERTa & \textbf{66} & 63.83 & \textbf{64} & 64.52 \\
        & Qwen-DepRoBERTa & \textbf{66} & 63.83 & 61 & 56.18 \\
        & GPT-Suicidality-ELECTRA & 63 & 64.76 & 60 & \textbf{69.70} \\
        & Qwen-Suicidality-ELECTRA & 60 & \textbf{67.74} & 55 & 58.72 \\
        & GPT-Suicide-RoBERTa & 58 & 63.16 & 53 & 66.67 \\
        & Qwen-Suicide-RoBERTa & 57 & 49.41 & 50 & 66.67 \\
        \bottomrule
    \end{tabular}}
    \vspace{-5mm}
    \label{tab:tab3}
\end{table}
\subsection{Text Modality}
The experimental results in Table~\ref{tab:tab3} demonstrate the dev set performance of various pretrained language models on the suicidal risk assessment on ER and ED tasks. For the Chinese models on the ER task, while directly fine-tuning the Chinese-BERT model on ASR results (\textbf{CBERT}) achieved Acc (57\%) and F1 (64.46\%), the \textbf{GPT-CBERT} model obtained the highest Acc (73\%) and F1 (75.23\%), demonstrating that fine-tuning \textbf{CBERT} with suicidal risk-related text features extracted by GPT-4o remarkably boosts its capability for predicting suicidal risk in Chinese text. For the ED task, \textbf{Qwen-CBERT} achieved the best performance with 59\% Acc and 65.55\% F1, suggesting that Qwen-plus based suicidal risk-related features are more effective in capturing ED task. 

For the English models, DepRoBERTa trained on direct English translations of ASR results achieved in ER (55\% Acc, 57.14\% F1) and ED (56\% Acc, 57.69\% F1). \textbf{GPT-DepRoBERTa} outperformed other comparisons in both ER (66\% Acc, 63.83\% F1) and ED (64\% Acc, 64.52\% F1) tasks, indicating that fine-tuning with GPT-4o-based suicidal risk-related features is more effective than using raw ASR results and boosts suicidal risk detection performance. Furthermore, \textbf{GPT-Suicidality-ELECTRA} achieved the highest F1 (69.70\%) for the ED task. A notable trend observed in both Chinese and English models is that fine-tuning with GPT-4o-based suicidal risk-related features consistently outperforms Qwen-based features in ER tasks, whereas Qwen-based outputs perform better in ED tasks. This suggests that different LLMs may be capturing distinct linguistic and emotional patterns relevant to suicidal risk assessment. 

Our key finding is that models trained on English suicidal risk-related features extracted from Chinese speech transcripts outperform those trained on Chinese features in the ED task. This result highlights the potential of leveraging LLMs for cross-linguistic suicidal risk feature extraction. Rather than relying solely on monolingual fine-tuning, our framework demonstrates that suicidal risk indicators extracted in one language (Chinese) can be effectively transferred and utilized in another language (English) without performance degradation.

\begin{table}[t!]
    \centering
    \caption{Dev set performance of different speech foundation models on suicidal risk detection. Acc (\%) and F1 (\%) are reported for both \textbf{ER}, \textbf{Passage Reading (PR)}, and \textbf{ED}. The best-performing models for each metric are highlighted in \textbf{bold}.}
    \renewcommand{\arraystretch}{1}
    \addtolength{\tabcolsep}{1pt}
    \resizebox{\linewidth}{!}{
    \begin{tabular}{l|cc|cc|cc}
        \toprule
        \multirow{2}{*}{\textbf{Method}} & \multicolumn{2}{c|}{\textbf{ER}} & \multicolumn{2}{c|}{\textbf{PR}} & \multicolumn{2}{c}{\textbf{ED}} \\
        & \textbf{Acc} & \textbf{F1} & \textbf{Acc} & \textbf{F1} & \textbf{Acc} & \textbf{F1} \\
        \hline
        \midrule
        
        WavLM & \textbf{61.48} & 50.59 & \textbf{59.04} & 60.15 & 59.83 & 61.29 \\
        XLSR & 56.38 & \textbf{58.04} & 57.03 & \textbf{64.99} & \textbf{61.92} & \textbf{65.40} \\
        wav2vec2-Chinese & 50.58 & 39.32 & 55.49 & 47.64 & 59.41 & 61.96 \\
        \bottomrule
    \end{tabular}}
    \vspace{-3mm}
    \label{tab:tab4}
\end{table}
\subsection{Speech Modality}
The results in Table~\ref{tab:tab4} show that \textbf{XLSR} outperforms \textbf{WavLM} and \textbf{wav2vec2-Chinese} in the PR and ED tasks, achieving the highest F1 (64.99\% for PR, 65.40\% for ED). Notably, \textbf{XLSR} surpasses the Chinese-specific wav2vec2 model, suggesting that cross-linguistic speech representations generalize well for suicidal risk detection. 

\subsection{Multi Modality and Final Prediction.}
\begin{table}[t!]
    \centering
    \caption{Dev set performance of multimodal approach (speech + text) for suicidal risk assessment. Acc (\%) and F1 (\%) are reported for \textbf{ER} and \textbf{ED}. \textbf{Combined} refers to the final prediction derived by combining results from the three tasks. The best-performing models for each metric are highlighted in \textbf{bold}.}
    \renewcommand{\arraystretch}{1}
    \addtolength{\tabcolsep}{1pt}
    \resizebox{\linewidth}{!}{
    \begin{tabular}{l|cc|cc|cc}
        \toprule
        \multirow{2}{*}{\textbf{Method}} & \multicolumn{2}{c|}{\textbf{ER}} &  \multicolumn{2}{c|}{\textbf{ED}} & \multicolumn{2}{c}{\textbf{Combined}} \\
        & \textbf{Acc} & \textbf{F1} & \textbf{Acc} & \textbf{F1} & \textbf{Acc} & \textbf{F1} \\
        \hline
        \midrule
        
        eGeMAPs+SVM (baseline) & 52 & - & 58 & - & 53 & - \\
        W2V2+BERT (baseline) & 56 & - & 54 & - & 56 & - \\
        GPT-CBERT+XLSR & 65 & \textbf{69.57} & \textbf{62} & 48.65 & 65 & 69.03 \\
        Qwen-CBERT+XLSR & \textbf{69} & 68.69 & 61 & \textbf{68.29} & \textbf{68} & \textbf{69.23} \\
        GPT-DepRoBERTa+XLSR & 59 & 57.73 & 61 & 64.22 & 61 & 61.39 \\
        Qwen-DepRoBERTa+XLSR & 59 & 57.73 & 61 & 64.22 & 61 & 61.39 \\

        \bottomrule
    \end{tabular}}
    \vspace{-3mm}
    \label{tab:tab5}
\end{table}
The multimodal results in Table~\ref{tab:tab5} confirm that integrating speech and text improves suicidal risk detection compared to the two baselines. \textbf{Qwen-CBERT+XLSR} achieves the highest Acc (69\%) in the ER task and the best F1 (68.29\%) in the ED task. \textbf{GPT-CBERT+XLSR} obtains the highest F1 (69.57\%) in ER but demonstrates a lower F1 (48.65\%) in ED. \textbf{Qwen-CBERT+XLSR} achieves the best performance in the combined results, which implies that our proposed method notably outperforms the baseline and highlights the effectiveness of using the suicidal risk-related features extracted by LLMs.

\subsection{Leaderboard Results}
Table~\ref{tab:tab6} presents both dev and test set results. Due to limited submission opportunities, we report only the results for \textbf{Qwen-CBERT+XLSR} and \textbf{GPT-DepRoBERTa+XLSR}. \textbf{W2V2+BERT} achieves the highest test set performance (61\%), outperforming all other models, despite having a moderate dev score (56\%).
\textbf{Qwen-CBERT+XLSR} performs best on the dev set (68\%) but drops significantly on the test set (54\%), suggesting potential overfitting or domain shift issues.
\textbf{GPT-DepRoBERTa+XLSR} shows a similar trend, indicating reduced generalization.

\begin{table}[t!]
    \centering
    \caption{Dev and test set performance of various models. The highest Acc (\%) and F1 (\%) is highlighted in \textbf{bold}.}
    \renewcommand{\arraystretch}{1}
    \addtolength{\tabcolsep}{1pt}
    \resizebox{\linewidth}{!}{
    \begin{tabular}{l|cc|cc}
        \toprule
        \multirow{2}{*}{\textbf{Method}} & \multicolumn{2}{c|}{\textbf{Dev}} &  \multicolumn{2}{c}{\textbf{Test}} \\
        & \textbf{Acc} & \textbf{F1} & \textbf{Acc} & \textbf{F1} \\
        
        \hline
        \midrule
        
        eGeMAPs+SVM (baseline) & 53 & - & 51 & - \\
        W2V2+BERT (baseline) & 56 & - & \textbf{61} & - \\
        Qwen-CBERT+XLSR & \textbf{68} & \textbf{69.23} & 54 & 54.90\\
        GPT-DepRoBERTa+XLSR & 61 & 61.39 & 47 & 47.52 \\

        \bottomrule
    \end{tabular}}
    \vspace{-5mm}
    \label{tab:tab6}
\end{table}

\section{Discussion}

\subsection{Limitation Statement}
The findings presented in this study are based on the scoring framework of the MINI-KID scale, which assesses current suicide risk as at risk or no risk. This classification reflects participants’ immediate responses to the MINI-KID assessment and should not be interpreted as a prediction of future suicidal behavior. Although the MINI-KID suicide module is widely recognized as a gold standard for assessing current suicide risk among adolescents, it has limitations. It relies heavily on self-reported data, which may lead to underreporting or misinterpretation of symptoms, and its fixed set of items may not fully capture the complex and dynamic nature of suicidal ideation and behavior. Accordingly, the results reported herein are strictly confined to the context of this assessment.

\section{Conclusion}
This study introduced a language-agnostic suicidal risk detection framework leveraging LLMs. 
Our method achieved performance comparable to or better than direct fine-tuning with ASR results or monolingual models trained exclusively on Chinese suicidal risk-related features.
However, it exhibited a significant performance gap between the development and test sets, suggesting a possible data distribution inconsistency between the two sets.
Future work can be considered on refining prompt strategies, transitioning from first-person perspective prompts to general summarization prompts as illustrated in~\cite{Sadeghi2024HarnessingMA}, enhancing generalization, and improving multilingual adaptability, contributing to scalable AI-driven mental health assessment and suicidal risk detection. 




\section{Acknowledgement}
This research was supported by a grant of the Korea Health Technology R\&D Project through the Korea Health Industry Development Institute (KHIDI), funded by the Ministry of Health \& Welfare, Republic of Korea (grant number: HI22C1962), and by Brian Impact Foundation, a non-profit organization dedicated to the advancement of science and technology for all.


\bibliographystyle{IEEEtran}
\bibliography{mybib}

\end{document}